\documentclass[letterpaper, 10 pt, conference]{ieeeconf}  

\IEEEoverridecommandlockouts                              

\usepackage{graphics} 
\usepackage{epsfig} 
\usepackage{amsmath} 
\usepackage{amssymb}  
\usepackage[ruled,linesnumbered]{algorithm2e}
\usepackage{multirow}
\usepackage{leftidx}
\usepackage{siunitx}

\usepackage{color} 
\usepackage{multicol}
\usepackage[normalem]{ulem}

\usepackage{draftwatermark}

\usepackage{tikz}
\usepackage{textcomp}
\usepackage{lipsum}

\newcommand\copyrighttext{%
  \footnotesize \textcopyright 2012 IEEE. Personal use of this material is permitted.
  Permission from IEEE must be obtained for all other uses, in any current or future
  media, including reprinting/republishing this material for advertising or promotional
  purposes, creating new collective works, for resale or redistribution to servers or
  lists, or reuse of any copyrighted component of this work in other works.}
 \newcommand\copyrightnotice{%
 \begin{tikzpicture}[remember picture,overlay]
 \node[anchor=south,yshift=10pt] at (current page.south) {\fbox{\parbox{\dimexpr\textwidth-\fboxsep-\fboxrule\relax}{\copyrighttext}}};
 \end{tikzpicture}%
 }
 
\title{\LARGE \bf
FBG-Based Position Estimation of Highly Deformable Continuum Manipulators: Model-Dependent vs. Data-Driven Approaches
}

\author{Shahriar Sefati$^{1}$, {\it Member, IEEE}, Rachel Hegeman$^{1,2}$, Farshid Alambeigi$^{1}$, {\it Member, IEEE},\\ Iulian Iordachita$^{1}$, {\it Senior Member, IEEE}, and Mehran Armand$^{1,2}$, {\it Member, IEEE}
\thanks{*Research supported by NIH/NIBIB grant R01EB016703 and Johns Hopkins internal funds.}
\thanks{$^{1}$S. Sefati, R. Hegeman, F. Alambeigi, I. Iordachita, and M. Armand are with the Laboratory for Computational Sensing and Robotics, Johns Hopkins University, Baltimore, MD, USA (sefati@jhu.edu, rachel.hegeman@jhuapl.edu ,falambe1@jhu.edu, iordachita@jhu.edu, mehran.armand@jhuapl.edu).
}%
\thanks{$^{2}$R. Hegeman, and M. Armand are with the Johns Hopkins University Applied Physics Laboratory, Laurel, MD, USA (rachel.hegeman@jhuapl.edu, mehran.armand@jhuapl.edu).
}%
}

\begin{document}

\SetWatermarkText{Accepted for ISMR 2019}
\SetWatermarkScale{0.4}

\maketitle
\thispagestyle{empty}
\pagestyle{empty}
\copyrightnotice

\begin{abstract}

Conventional shape sensing techniques using Fiber Bragg Grating (FBG) involve finding the curvature at discrete FBG active areas and integrating curvature over the length of the continuum dexterous manipulator (CDM) for tip position estimation (TPE). However, due to limited number of sensing locations and many geometrical assumptions, these methods are prone to large error propagation especially when the CDM undergoes large deflections. In this paper, we study the complications of using the conventional TPE methods that are dependent on sensor model and propose a new data-driven method that overcomes these challenges. The proposed method consists of a regression model that takes FBG wavelength raw data as input and directly estimates the CDM's tip position. This model is pre-operatively (off-line) trained on position information from optical trackers/cameras (as the ground truth) and it intra-operatively (on-line) estimates CDM tip position using only the FBG wavelength data. The method's performance is evaluated on a CDM developed for orthopedic applications, and the results are compared to conventional model-dependent methods during large deflection bendings. Mean absolute TPE error (and standard deviation) of $1.52$ ($0.67$) $mm$ and $0.11$ ($0.1$) $mm$ with maximum absolute errors of $3.63$ $mm$ and $0.62$ $mm$ for the conventional and the proposed data-driven techniques were obtained, respectively. These results demonstrate a significant out-performance of the proposed data-driven approach versus the conventional estimation technique.   

\end{abstract}


\section{INTRODUCTION} \label{intro}

Due to their flexibility, continuum dexterous manipulators (CDMs) can be used in surgical tasks where significant dexterity and maneuverability are necessary. Compared to rigid link robots, CDMs can adopt various shapes, therefore, enhance the reach of surgeons in confined spaces \cite{walker2016snake}. Though beneficial for complex surgical tasks, a CDM's flexibility makes accurate shape sensing and TPE challenging. Further, in many applications CDMs (usually the tip) are guided to target locations inside the body, requiring accurate position sensing for safe navigation.

In recent years, optical fibers such as Fiber Bragg Gratings (FBG) have been the focus of attention for shape and tip sensing of biopsy needles (e.g.\cite{roesthuis2014three}) and continuum manipulators \cite{sensingTechniques}. FBGs are flexible, small in size and can provide data at high frequencies without requiring a direct line of sight making them perfectly suitable for CDM position sensing, especially for surgical applications. The fibers are typically either combined with small elastic substrates to form a standalone sensor \cite{liu2015shape} or attached directly to the CDMs \cite{xu2016curvature}.

\begin{figure}[!t] 
	\centering
	\includegraphics[width=\linewidth]{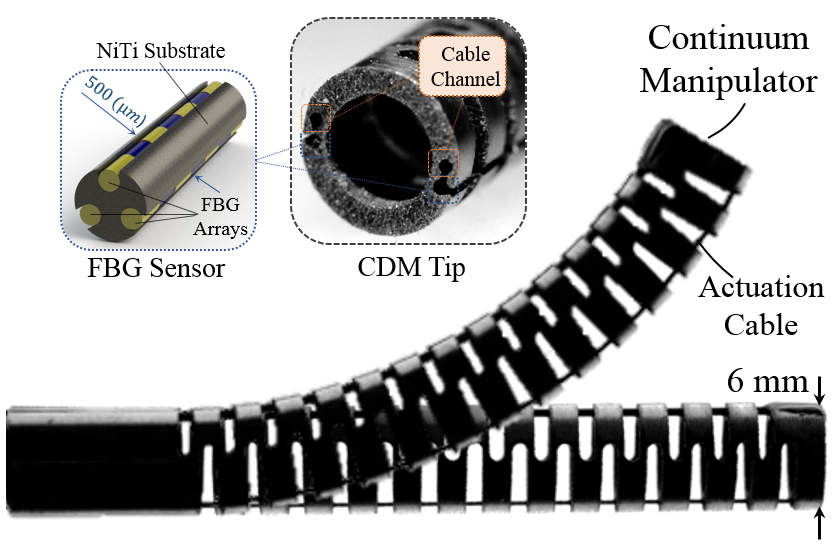}
	\caption{Continuum manipulator developed for orthopedic applications as well as the integrated FBG sensor with triangular configuration. The sensor consists of three fibers, each with three active areas.}
	\label{fig:snake}
\end{figure}

A variety of FBG sensor manufacturing techniques associated with the choice of substrate material, geometry, number of fibers and their attachments have been explored in the literature 
\cite{liu2015shape}-\cite{sefatiVibration}.
Liu et al. \cite{liu2015shape} used a triangular configuration (Fig. \ref{fig:sensor_assembly}) with one fiber (three gratings) and two Nitinol (NiTi) substrates. They first found the curvature at discrete locations by calibrating the FBG wavelength data to constant curvatures and then performing TPE indirectly using this calibration. Roesthuis et al. \cite{roesthuis2013CDM} attached three fibers each with four active areas to a NiTi substrate and found curvature and bending plane by modeling the sensor assembly's cross section. Xu et al. \cite{xu2016curvature} proposed a helically wrapped sensor with three FBG fibers for simultaneous curvature/torsion/force measurements for concentric tube robots. Wei et al. \cite{wei2017} also suggested a helically wrapped sensor using only one fiber with multiple FBG nodes.

A common technique for FBG-based TPE is to find curvature at discrete locations, extrapolate curvature along the length of the sensor, compute slopes and reconstruct the shape from base to tip  \cite{sensingTechniques}. In a triangular configuration (Fig. \ref{fig:sensor_assembly}), three FBG nodes at each cross sections are used to simultaneously find curvature, plane of bend, and the common strain at active sensing locations \cite{roesthuis2014three}. This method of TPE, however, relies on many geometrical assumptions (model-dependent) contributing to inaccuracies in estimations of curvature and plane of bend. Moreover, due to the limited number of active sensing locations and shape reconstruction from base to tip, the TPE will be prone to error, especially in large deflections. 

To address the mentioned drawbacks in conventional model-dependent shape reconstruction and TPE techniques, we propose a new data-driven approach that uses only the sensory data to learn the necessary parameters for estimating the CDM's tip position. The trained model does not rely on any assumption regarding the geometrical relations of the manufactured sensor and is therefore expected to reduce the error in CDM TPE especially when CDM undergoes large deflections. The contributions of this paper are as follows: 1) proposing a data-driven model-independent approach using only sensory data to estimate CDM tip position; 2) implementation and analysis of both the proposed and the conventional TPE methods on a large deflection CDM developed for orthopedic applications; 3) error analysis of the conventional model-dependent technique on large deflection CDMs; and 4) comparison of TPE error using different number of fibers (one, two, or three) in the data-driven approach.

The remainder of the paper is organized as follows. Section \ref{methods} reviews the conventional model-dependent approach and describes the proposed data-driven method for CDM TPE. Section \ref{experiments} explains the CDM and FBG sensor used in this study, and describes the experiments conducted for evaluation of the proposed and conventional methods. Section \ref{results} includes the TPE results and discussion on the obtained data from the experiments. In section \ref{conclusion} the paper is concluded and future work is presented.

\section{METHODS} \label{methods}

As the CDM bends, the raw FBG wavelength data changes due to strain changes at the FBG nodes. Conventional shape reconstruction methods \cite{sensingTechniques}
-\cite{roesthuis2013CDM} aim at estimating curvature at active sensing locations based on the sensor's geometrical model and then integrating over the sensor's length to estimate the tip position. A few drawbacks of this method are 1) curvature is found at each cross section using only the local FBG nodes in that specific cross section 2) at minimum, three FBG nodes are required at each active cross section 3) several assumptions regarding the sensor design are made during the shape reconstruction step 4) extra calibration steps are required to find geometrical parameters. To remedy the above-mentioned shortcomings, we propose a data-driven approach where the correlation of all FBG nodes on the sensor is used simultaneously to predict the CDM tip position. In addition, the data-driven approach does not make any geometrical or structural assumptions regarding the sensor design.

\subsection{Conventional Model-dependent Method} \label{methods:model_based}

\begin{figure}[!t] 
	\centering
	\includegraphics[width=\linewidth]{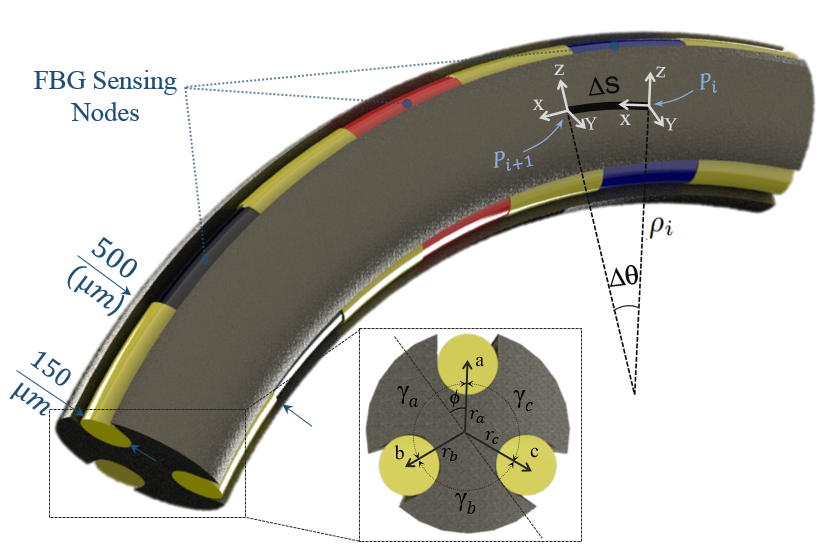}
	\caption{FBG sensor and its cross section in triangular configuration. Geometrical parameters used in conventional model-dependent approach are shown.}
	\label{fig:sensor_assembly}
\end{figure}

In model-dependent approaches, first the strain at each FBG node is found by (\ref{eq:epsilon}) and then based on the sensor's geometry, a system of nonlinear equations (\ref{eq:systemOfEqs}) is solved at each cross section with three FBG nodes (fibers a, b, and c placed in a triangular configuration (Fig. \ref{fig:sensor_assembly}) to find curvature ($\kappa$), its angle ($\phi$), and a common strain bias ($\epsilon_0$) at that cross section of the sensor:

\begin{equation} \label{eq:epsilon}
\epsilon = \frac{\Delta \lambda_B}{\lambda_B (1-p_e)}
\end{equation}

\begin{equation} \label{eq:systemOfEqs}
\begin{split}
\epsilon_a &= -\kappa r_a sin(\phi) + \epsilon_0 \\
\epsilon_b &= -\kappa r_b sin(\phi + \gamma_a) + \epsilon_0 \\
\epsilon_c &= -\kappa r_c sin(\phi + \gamma_a + \gamma_b) + \epsilon_0
\end{split}
\end{equation}
where $\lambda_B$ is the Bragg wavelength, $p_e$ is the strain constant for optical fiber and $r$, $\gamma$ are geometrical parameters and can be assumed known from design or be found by an extra calibration procedure \cite{roesthuis2014three}. Then assuming a relationship (typically linear) between curvature and arc length, and dividing the sensor length to $n$ sufficiently small segments, curvature ($\kappa$) and its direction ($\phi$) can be extrapolated at each segment (\ref{eq:curv-arc}). Using the curvature at each segment ($\kappa_i$ for $i=1,...,n$) slope of each segment can be found using (\ref{eq:slope-curve}). By attaching an appropriate local coordinate frame to the beginning of each segment, shape can be reconstructed segment by segment to find the CDM tip position using (\ref{eq:3D-pos}):

\begin{equation} \label{eq:curv-arc}
\kappa = f(s) , \phi = g(s)
\end{equation}
\begin{equation}\label{eq:slope-curve}
\Delta \theta_i = \frac{\Delta s}{\rho_i} = \kappa_i \Delta s
\end{equation}
\begin{equation} \label{eq:3D-pos}
\begin{gathered}
P_{i+1} = R_{i}P_{i} + \Delta P_{i} \\
R_{i} =
\begin{bmatrix}
1 & 0 & 0 \\
0 & cos(\phi) & -sin(\phi)\\
0 & sin(\phi) & cos(\phi)\\
\end{bmatrix} \\
\Delta P{i} =
\begin{bmatrix}
\rho_{i}sin(\Delta \theta_{i}) &
0 &
- \rho_{i} (\rho_{i} - \rho_{i}cos(\Delta \theta_{i})) 
\end{bmatrix}^T
\end{gathered}
\end{equation}

Using (\ref{eq:3D-pos}) for $i= 1,...,n$, the tip position of the sensor ($P_{n}$) is found. For CDMs in which the FBG sensor is not placed on the center axis of the CDM, all $P_{i}$s should then be shifted by $L_{shift}$ (distance between sensor's and CDM's centers) to obtain the 3-D position of the center-line of the CDM.

\subsection{Data-driven Method} \label{methods:data_driven}
To improve the accuracy of TPE, we propose a data-driven method that does not rely on any assumptions regarding the design and manufacturing of the sensor or the number of FBG nodes at each cross section. In addition, the proposed method simultaneously uses data from all FBG nodes across the sensor length for TPE. 

FBG is a suitable sensing modality during a surgery where there is no or limited direct line of sight. Moreover, it can stream data at high frequencies (e.g. $1$ KHz), which is ideal for real-time CDM tip position control. Other sensing modalities such as optical trackers or stereo cameras can stream the data in lower frequencies, however, can measure the CDM tip position directly with high accuracy. Therefore, a model can be trained on the FBG raw wavelength data as input and the true CDM tip position from another sensor (optical tracker or cameras) as output. The training phase can be carried out pre-operatively (off-line) and the trained model can be used to predict the CDM tip position intra-operatively using only the FBG raw wavelength data as input. A regression model is suitable for the proposed task, since given some independent variables as input, prediction of a dependant variable is desired.

For the problem at hand, we aim to find a regression model that can predict the CDM 3-D tip position (dependent variables) given a set of raw FBG wavelength data along the sensor (independent variables):

\begin{equation}
p = \Psi(\lambda, \beta)
\end{equation}
where $p \in \mathbb{R}^{3}$ is the 3-D position of the CDM tip, $\lambda \in \mathbb{R}^{m}$ is the vector containing the raw wavelength data of the \textit{m} FBG nodes on the sensor, $\beta$ is the vector of unknown parameters, and $\Psi: \mathbb{R}^{m} \rightarrow \mathbb{R}^3$ is the regression model that predicts CDM 3-D tip position, given the wavelength information of the complete set of FBG nodes on the sensor at any given time. Depending on the degrees of freedom of the CDM, complexity of the environment, and the shapes that the CDM can obtain, different regression models could be used to capture the unknown parameters $\beta$. Linear regression is a common regression model that models the dependent variables as a linear combination of the unknown parameters. The CDM tip position sensing can be modeled as a least squares optimization problem:

\begin{equation} \label{eq:regression}
\begin{aligned}
& \underset{B} {\text{argmin}}
& & \sum_{n=1}^{N} r_n^2
=
& \underset{B} {\text{argmin}}
& & \| \Lambda B - P \|^2_2 
\end{aligned}
\end{equation}
where $r_n$ is the residual of the error for the $n^{th}$ observation, $\Lambda \in \mathbb{R}^{N \times m}$ is a stack of \textit{N} observations of the \textit{m} FBG node data, $P \in \mathbb{R}^{N \times 3}$ is the stack of \textit{N} true CDM tip position observation data from optical tracker or cameras, and $B \in \mathbb{R}^{m \times 3}$ is the matrix of unknown parameters. Using (\ref{eq:regression}), the regression model is trained pre-operatively on N observations of the FBG and true CDM tip position (section \ref{exp:optical_tracker}) to find the unknown parameters $B$ using the generalized inverse:

\begin{equation} \label{eq:unknown-params}
B = (\Lambda^T \Lambda)^{-1} \Lambda^T P
\end{equation}

The trained model can then be used intra-operatively to predict CDM tip position values given only the current FBG wavelength data:

\begin{equation} \label{eq:prediction}
\hat{p} = B^T \lambda
\end{equation}
where $\hat{p} \in \mathbb{R}^3$ is the predicted CDM tip position, given the current wavelength data for all FBG nodes on the sensor ($\lambda \in \mathbb{R}^m$).

Using the proposed regression model, the uncertainties regarding the shape sensor manufacturing such as FBG locations will be captured in the unknown parameters matrix $B$.  In addition, the wavelength information of all FBG nodes (and not just the ones at a certain cross section) on the sensor are utilized for tip position prediction. Further, the regression model does not suffer from the accumulated errors that occur during integration of position over the length of the CDM in conventional methods, since the tip position is predicted directly. Moreover, the regression model can be trained separately on the data from fewer than three FBG fibers, to still be able to predict the CDM tip position in case of fiber failure.

\begin{figure*}[!t] 
	\centering
	\includegraphics[width=\linewidth]{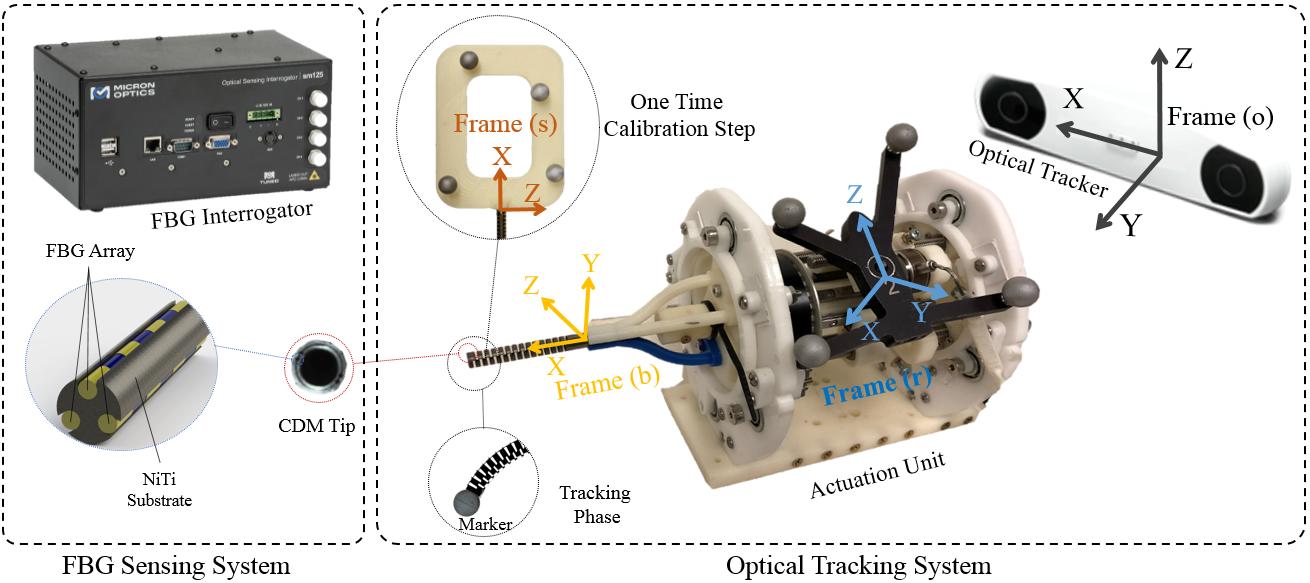}
	\caption{Experimental setup including an actuation unit and an FBG interrogator. Coordinate frames are shown: optical tracker (o), reference body (r), CDM straight pose body (s), and the base of the CDM (b).}
	\label{fig:setup}
\end{figure*}

\section{Experimental Setup and True Tip Estimation} \label{experiments}

\subsection{CDM and FBG Sensor Design} \label{exp:CDM-FBG}
The CDM used for the experiments was previously developed for orthopedic applications such as the treatment of osteolysis \cite{alambeigi2016design}, \cite{alambeigi2018acc} or osteonecrosis \cite{alambeigi2017core} (Fig. \ref{fig:snake}). It is constructed from a NiTi tube with several notches to enable a flexible bend. The outside diameter (OD) of the CDM is $6$ mm, with an inner $4$ mm open lumen for passing tools such as those used for debridement \cite{alambeigi2016design}. The CDM's thin wall contains channels for passing the actuation cables and the shape sensing units. The FBG sensor used for this study has three FBG fibers attached to a flexible NiTi wire with an OD of $0.5$ mm in a triangular configuration (Fig. \ref{fig:sensor_assembly}). Three grooves (radially \ang{120} apart from each other) are engraved by laser (Potomac, USA) along the length of the wire to hold three fibers each with three FBG nodes (Technica S.A, China). Due to its relatively small OD, the NiTi wire can withstand curvatures of as small a radius as $20$ mm during bending, which is sufficient to achieve large deflections of the CDM \cite{sefati2017fbg}.
   
The CDM cables are actuated with two DC motors (RE16, Maxon Motor Inc. Switzerland) with spindle drives (GP 16 A, Maxon Motor, Inc. Switzerland) on the actuation unit shown in Fig. \ref{fig:setup}. A commercial controller (EPOS 2, Maxon Motor Inc. Switzerland) is used to control the individual Maxon motors. Using libraries provided by Maxon, a custom C++ interface performs independent velocity or position control of each motor of the CDM's actuation unit. 

\subsection{CDM Tip Sensing by Optical Trackers} \label{exp:optical_tracker}

Optical trackers can provide 3-D position information of the reflective markers with  RMS error ranging between $0.25$ and $0.35$ mm \cite{polaris} with relatively high frequencies up to $60$ Hz, however, they are not suitable for tracking the tip position of the CDM intra-operatively since they require a direct line of sight. Consequently, we use an optical tracker (Polaris, NDI Inc.) for true CDM tip position during the pre-operative (off-line) regression model training, even though other sensors such as stereo cameras, etc can also be used for this purpose.

We follow a two step procedure to find the tip position of the CDM relative to its base using the optical tracker (frame \textit{o}). First, we establish a coordinate frame (\textit{s}) at the CDM tip when it is straight with respect to a reference body that is attached to the fixed actuation unit (frame \textit{r}) using a 3-D printed rigid body placed into the open lumen of the CDM (Fig. \ref{fig:setup}):

\begin{equation}
T_{sr}^{0} = (T_{os}^{0})^{-1} \: T_{or}^{0}
\end{equation}
where $T_{ij}^{0}$ denotes frame \textit{j} with respect to frame \textit{i} during the one-time coordinate frame establishment (time $0$). The advantage of knowing the transformation between frames \textit{s} and \textit{r} is that later all the information will be reported with respect to the reference body, independent of where the optical tracker is located.

For tip tracking during CDM bending, a single spherical marker attached to a post is placed into the instrument channel of the CDM. The tracked position with respect to the optical tracker coordinate ($P_o$) frame can be represented with respect to the base of the CDM (frame \textit{b}) by:

\begin{equation} \label{eq:trackerPos}
P_b = T_{bs} \: T_{sr}^{0} \: (T_{or}^{c})^{-1} \: P_o
\end{equation}
where $T_{or}^{c}$ is the reference body coordinate frame with respect to the optical tracker coordinate frame at its current location (in case the location of optical tracker changes). $T_{bs}$, brings the CDM tip position explained in frame \textit{s} to the CDM base frame ($P_b$):

\begin{equation}
T_{bs} = 
\left[
\begin{array}{c|c}
I_{3 \times 3} & P_{bs} \\
\hline
0_{1 \times 3}& 1
\end{array}
\right]
\end{equation}
where $P_{bs} = [0, 0, L_{CDM}]^T$ and $L_{CDM}$ is the length of the CDM. From Eq. \ref{eq:trackerPos}, the CDM tip positions is known with respect to the coordinate frame at the base of the CDM (frame \textit{b}).

\section{Evaluation Experiments} \label{experiments-design}

The CDM used in this study is designed with large deflection capabilities for enhanced dexterity for orthopedic applications. Therefore, it is essential that the tip position of the CDM is estimated accurately during large deflections, since this estimation is used as feedback for position control of the CDM's tip during surgery \cite{sefati2018IROS}.

To evaluate the proposed data-driven method (section \ref{methods:data_driven}), the CDM was bent from straight pose to large tip deflections of $22$ mm (corresponding to $62\%$ of CDM's length and $50 [m]^{-1}$ curvature) and the sensor's raw FBG wavelength data as well as the true tip position from the optical tracker were collected. The collected data set was then randomly split to $70\%$ training and $30\%$ testing data and this random split was repeated $10$ times to ensure the model's performance is not dependant on the choice of the training data. The regression model was trained on the training data and evaluated on the unseen test data. In addition, the large deflection bending experiment was repeated $5$ times to evaluate the repeatability of the CDM tip prediction via the regression model. The same data set was used to predict the CDM's tip position via the conventional model-dependent approach (section \ref{methods:model_based}) and the results were compared.

During the experiments, the CDM's cables were actuated with a velocity of $0.05$ mm/s to obtain data with good resolution. FBG data was streamed by an optical sensing interrogator (Micron Optics sm 130) at $100$ Hz frequency, while optical tracker data (Polaris Vicra, Northern Digital Inc., Canada) was collected at $20$ Hz frequency. To this end, a thread-safe mechanism from open source c++ CISST-SAW libraries (https://github.com/jhu-cisst/cisst-saw) was used to record the data for each of these sensor sources in parallel, using a separate CPU thread for each source. All the data was time-stamped and pre-processed during the training phase to time-align the FBG and optical tracker data samples. All experiments were run on a computer with an Intel 2.3 GHz core i7 processor with 8GB RAM, running Ubuntu $14.04$.

   \begin{figure}[t] 
      \centering
      \includegraphics[width=\linewidth]{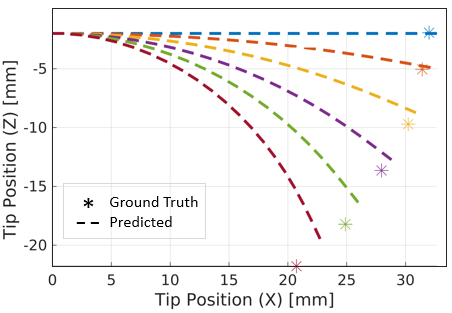}
      \caption{Shape reconstruction and TPE using the conventional model-dependent approach. Stars represent the ground truth data from the optical tracker.}
      \label{fig:model_results}
   \end{figure}

      \begin{figure}[t] 
      \centering
      \includegraphics[width=\linewidth]{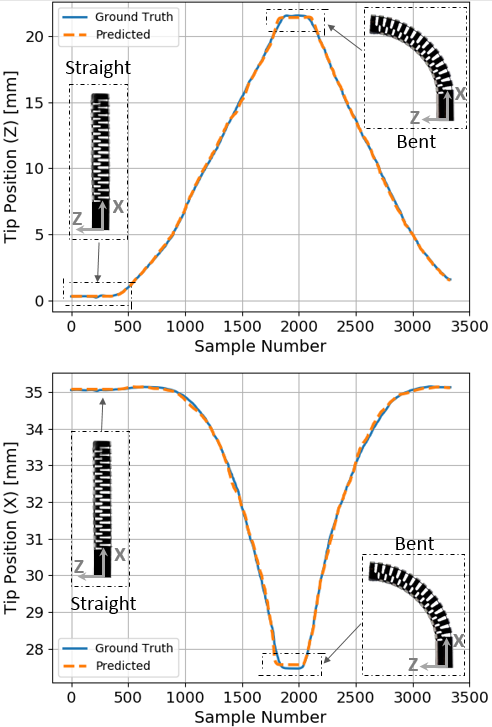}
      \caption{TPE in each Cartesian direction using the proposed data-driven approach on the test data-set that is unseen by the regression model during training. Blue and orange represent the ground truth and estimated CDM's tip position, respectively.}
      \label{fig:data_driven_results}
   \end{figure}

\section{Results and Discussion} \label{results}

Fig. \ref{fig:model_results} demonstrates the performance of the conventional model-dependent approaches in TPE of the CDM. In this figure, dashed lines show the CDM shape reconstructed, whereas stars represent the true tip CDM position from optical tracker data. In these experiments, the CDM was bent from a straight pose all the way to large curvatures of approximately $50$ $[m]^{-1}$. It is observed that the model-dependent approach can estimate the tip position relatively accurately for tip deflections below $10$ mm (corresponding to $28\%$ of the CDM's length) with mean absolute error (and standard deviation) of $0.7$ ($0.3$) mm. However, for larger deflections the deviation between the estimated and true tip position increases, exhibiting larger errors in TPE of the CDM. This increase of error with larger deflections may be due to imprecise assumptions on the sensor's and CDM's geometries, as well as error propagation due to reconstructing shape from base to tip in the conventional model-dependent method. In addition, due to limited number of active sensing locations along the sensor, even small errors in curvature estimation for active locations closer to the base of the CDM can result in large tip estimation errors. Using conventional model-dependent approach, tip estimation error mean (and standard deviation) of $1.9$ ($0.7$) mm and $1.52$ ($0.67$) mm are observed for CDM tip deflections above $10$ mm and for all deflections, respectively.    

As mentioned in section \ref{experiments-design}, the regression model is trained on a random split of the data-set ($70\%$ training data and $30\%$ testing data) with a training time of nearly $6$ seconds. Fig. \ref{fig:data_driven_results} demonstrates the performance of the trained model on the unseen test data-set using only the raw FBG data. The predicted tip position values in both directions of motion are shown in orange dashed lines while the ground truth tip position information from the optical tracker is displayed in blue solid line. It is observed that the tip position prediction from the regression model is following the true tip position in both directions and also for large deflections very accurately, with a mean (and standard deviation) of $0.11$ ($0.1$) mm for all deflections. Of note, to ensure the trained model's performance is not dependant on the specific choice of training data, the training and testing phases have been performed $10$ times with completely random split of the data-set, and similar performance is observed, with a $0.05$ mm deviation in tip position mean absolute error.


\begin{table}
\centering
\caption{cdm tpe error comparison between the conventional sensor model-dependant method and the data-driven model using different number of fibers.}
\label{table:results}
\begin{tabular}{c|c|c|c|}
\cline{2-4}
\multirow{2}{*}{}                                                                            & \multicolumn{3}{c|}{\begin{tabular}[c]{@{}c@{}}Manipulator Tip Position \\ Error [mm]\end{tabular}} \\ \cline{2-4}  & Mean  & Std. & Max \\ \hline
\multicolumn{1}{|c|}{\begin{tabular}[c]{@{}c@{}}Conventional Approach \\ (Three Fibers Required)\end{tabular}} & 1.52                          & 0.67                          & 3.63                          \\ \hline
\multicolumn{1}{|c|}{\begin{tabular}[c]{@{}c@{}}Data-driven Regression \\ (One Fiber Used)\end{tabular}}          & 1.98                          & 1.13                          & 4.62                           \\ \hline
\multicolumn{1}{|c|}{\begin{tabular}[c]{@{}c@{}}Data-driven Regression \\ (Two Fiber Used)\end{tabular}} & 0.3                          & 0.16                          & 1.01                          \\ \hline
\multicolumn{1}{|c|}{\begin{tabular}[c]{@{}c@{}}Data-driven Regression \\ (Three Fiber Used)\end{tabular}} & 0.11                          & 0.10                          & 0.62                          \\ \hline
\end{tabular}
\end{table}

One of the advantages of the proposed data-driven regression method is that the CDM tip position can still be estimated in 3-D with less than three FBG fibers at each cross section. This can be beneficial in case one or two fibers malfunction or get damaged during the surgery. Table \ref{table:results} summarizes the results obtained from TPE using the conventional model-dependent approach and the data-driven regression method using different number of FBG fibers. It can be concluded that the data-driven method not only improves performance compared to the conventional model-dependent method, but is still capable of estimating the tip position in absence of one or two FBG fibers as well.
                                  
\section{CONCLUSION} \label{conclusion}
In this study we proposed a data-driven regression model for TPE of a CDM with large deflections that did not require any prior assumptions regarding the sensor geometry (FBG node location and substrate geometry), and could still predict CDM tip position with fewer than three fibers. The regression model was trained in a short amount of time pre-operatively (off-line) on a randomly chosen data-set of FBG raw wavelength data as input and true tip position from optical tracker as output. It could then predict the CDM tip position intra-operatively (on-line) using only the FBG raw wavelength values from the sensor's current readings.

In this paper, performance of the proposed data-driven method was studied on a CDM developed primarily for orthopedic applications with large deflections. Results exhibited accuracy improvement especially in large deflections compared to the conventional model-dependent approach. This improvement could be explained as a result of making no assumptions regarding the sensor's geometry, as well as simultaneously using all FBG nodes on the sensor to estimate the CDM tip position, rather than only a few at specific cross sections. In addition, using the data-driven approach, even with fewer than three fibers the FBG sensor could still be functional in estimating the CDM tip, in case of fiber malfunction or damage during surgery.

It should be noted the proposed method is independent of the sensor modality (e.g. optical trackers or stereo cameras) used for off-line true CDM tip position measuring that is fed to the regression model during the training phase. Future work includes study of more complex CDMs and shapes, for which nonlinear regression models or different learning techniques may be required.






\end{document}